# CORRIDRONE: Corridors for Drones
## An Adaptive On-Demand Multi-Lane Design and Testbed

Lima Agnel Tony, Ashwini Ratnoo, and Debasish Ghose
Indian Institute of Science, Bengaluru 560012, India
Email: <limatony+ratnoo+dghose>@iisc.ac.in

In the past few decades, modes of travel and transport have evolved at a rapid pace. Unmanned Aerial Vehicles (UAVs) are the modes of transport that cater to many interesting applications [1, 2] and have made research in this field exciting. Introducing autonomous vehicles in Class-G airspace requires preparedness in terms of safety, security, and protection of human life and habitat, especially in urban scenarios. Many applications where UAVs are employed require airspace usage for a short span of time to move essential goods and material from one point to another. The applications demand a set-up that could be deployed almost anywhere with least effort and technical complexity. There is a necessity for a simple and easy to operate architecture, which could function with minimal hardware infrastructure and software.

In this article, a novel drone skyway framework called CORRIDRONE is proposed. As the name suggests, this represents virtual air corridors for point-to-point safe passage of multiple drones. The corridors are not permanent but can be set up on demand. An artists' impression of how these drone corridors will appear in the city of Mumbai is given in Fig. 1. A few such scenarios could be those in warehouse/factory floors, package delivery, shore-to-ship delivery, border patrol, etc. Several factors play major roles in the planning and design of such aerial passages. The proposed framework includes many novel features which aid safe and efficient integration of UAVs into the airspace with already available technologies. A several kilometres long test bed is proposed to be set-up at the 1500 acres Challekere campus of Indian Institute of Science, in the state of Karnataka, to design and test the infrastructure required for CORRIDRONE

**UAV traffic management systems**
Research till date has focused on integrating unmanned systems into the controlled airspace. The challenges of the problem are being crystalized and solved in steps. NASA UTM [3] and SESAR U-space [4] are among the well-known initiatives in this effort. A UAV traffic management system in the Indian context is proposed in [5]. There are many components that are essential in setting up a drone traffic management system. A few of these include path planning, geo-fencing, monitoring, traffic routing, communication, etc., which come under software requirements. The hardware requirements like drone ports for takeoff and landing, communication antennae set-up, the drone itself, equipped with appropriate sensors and components of desired capability, etc., makes the problem more demanding. Several promising results have been published for path planning in [6,7] and collision avoidance [8-11]. In [12] an optimization based multi-agent trajectory planning technique using quadratic programming is presented. Geo-fencing is the virtual barrier that prevents the UAVs from moving in/out of authorized air space. A thorough discussion on geo-fencing is presented in [13] and [14].

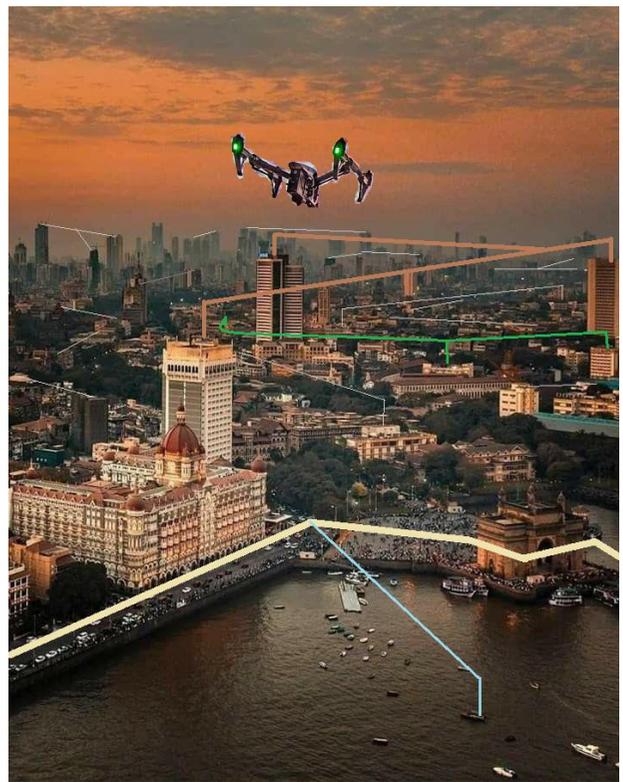

***Fig. 1*** *CORRIDRONE drone corridors envisaged in the Mumbai skyline*

However, most of the existing fencing techniques in the literature handle single lane traffic.

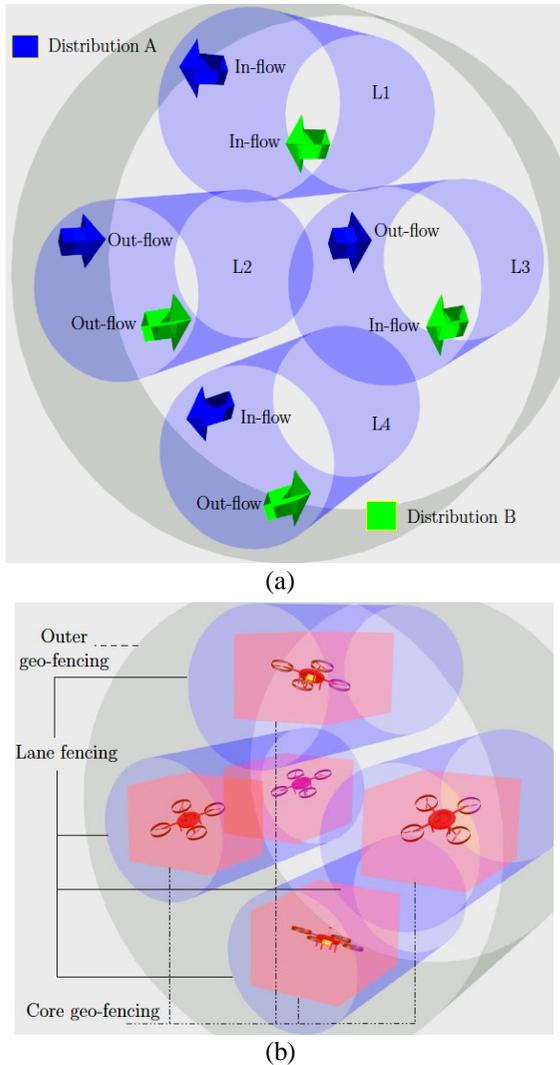

*Fig. 2 (a) Corridor cross section showing traffic flow pattern in a multi-lane design (b) Corridor cross section showing the different fencing layers in the proposed design*

The importance of communication and its application to the aerial traffic management system is discussed in [15]. A review of autonomous/semi-autonomous architectures is presented in [16], highlighting the gap in current technologies and the requirements of a fully autonomous system for UAV traffic management.

This article gives details of the setup of drone corridors, the aerial pathways for UAVs, which could be deployed with existing hardware and software technologies. The proposed framework demonstrates the possibility of employing drones for diverse applications without waiting for the development of a full-fledged and widely approved architecture. The CORRIDRONE discussed here is envisioned in a modular set-up in a way that it could be deployed anywhere with minimum hardware/software requirements and logistics; and could fit into any existing UTM system. This article also gives details of the proposed testbed being set up to develop and implement CORRIDRONE.

**CORRIDRONE: A multi-lane corridor for drones**

The CORRIDRONE is an end-to-end solution for setting up a drone passage on an ad hoc basis. This could be operated for various utilities which require only basic functionalities of UAV traffic management, for Beyond Visual Line-of-Sight (BVLOS) operations. The important features of the CORRIDRONE architecture are summarized below.

*1. Ground Control System (GCS)*

The proposed aerial corridor could be set up anywhere, given the permissions and related geographical information. The GCS and the launch/landing pads, are the minimum physical infrastructure requirements for the setup, other than the UAV. The software framework necessary for the design and planning of lanes and fencing for CORRIDRONE resides in the GCS. The GCS also monitors the vehicles' flying through the corridor at any given time, ensuring the safety and normal operation of the vehicles. It scans important parameters of the UAVs related to vehicle health, time of operation, location, etc., which helps the UAV operators take quick actions in any adversity. These functionalities of GCS make it portable for any on-demand application.

*2. Corridor geometry and lane planning*

An efficient corridor design is crucial for the smooth flow of traffic. The main corridor is defined first that joins the points between which the traffic must flow. The main corridor consists of multiple sub-corridors through which multiple drones could traverse at any given time (Fig. 2). The traffic flow patterns for lanes of multi-layer geo fence can be envisaged in many ways, two of which are shown in Fig. 2(a). Distribution A considers lanes L1, L4 for inflow, and L2, L3 for outflow which would facilitate better operation from the viewpoint of spatial separation of the lanes as well as being apt for takeoff and landing which may take place in mutually exclusive spaces. Also, lane changing needs maneuvers only in horizontal or vertical plane for this configuration. Distribution B has lanes L1, L3 designated for inflow while lanes L2, L4 provide outflow of UAVs. This configuration provides for lower aerodynamic downwash effects. The scenario is shown in Fig. 3, where various possibilities are shown. Lanes which are one over another contribute to vortex interaction between the drones, affecting the flight of the lower drone in terms of stability and higher control effort (Fig. 3(a)). Proper lane geometry (Fig. 3(b)) helps reduce it effects. Similar effects can be seen, if the drones are flying at the same speed in the same direction (Fig. 3(c)).

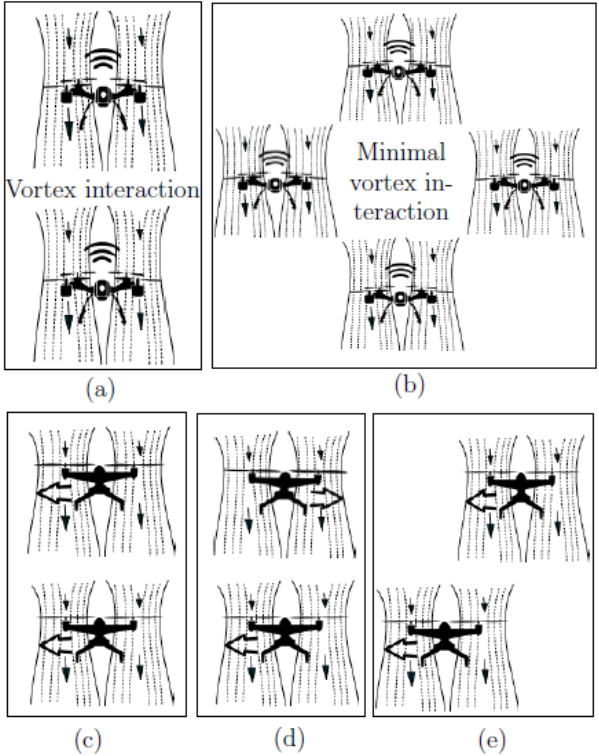

*Fig. 3 Effects of downwash (a) in stacked lanes (b) in planned lanes (c)-(e) while flying over one another*

The effects could be reduced if the configurations given in Fig. 3(d)-(e) can be achieved. In Fig. 3(d), due to opposite orientation of flight direction, minimal vertical interaction of drones can be ensured, while in Fig. 3(e), a stagger formation with same direction of flight will ensure the same, thus leading to less risk of inter-drone downwash interaction.

Exact usage of lanes would depend on the UAV throughput via CORRIDRONE. For example, using Distribution B with just two lanes L2 and L3 provides the basic CORRIDRONE configuration with minimum downwash interaction, and well-separated landing and take-off locations for moderate to low throughput. Spatio-temporal aspects of instantaneous UAV distribution across the corridor also take into account collision avoidance and geofencing constraints which are discussed subsequently.

*3. Adaptive dynamic geo-fencing*

Geo-fencing is a key component of the drone corridor design as it designates the strict boundaries of the lanes and sub-lanes for safety and smooth flow of drone traffic. The CORRIDRONE framework has a geo-fencing design as shown in Fig. 2(b). The outer fencing is the main corridor and lanes within are separated from each other using cylindrical geo-fences. Diameters of the corridor and lane geo-fences depend primarily on available airspace, size of UAVs, and dynamic constraints [17] of the UAVs. These geo-fences are adaptive and can be created on-demand for setting up the corridor. The basic geo-fencing structure ensures that lanes do not intersect or occupy no-fly regions.

In addition, each UAV also has a core geo-fence which is vital for safe operations in the presence of uncertainties and conflicts. Fig. 4 shows the core geo-fence as a cuboidal volume around a UAV moving through the corridor. As shown in the figure, the core geo-fence is defined using the total length $d_t$, the forward clearance $d_f$ along direction of motion, and rear clearance $d_f$. The total length $d_t$ of the core geo-fence is decided using the compliance level of the UAV, further details of which are discussed later in this section.

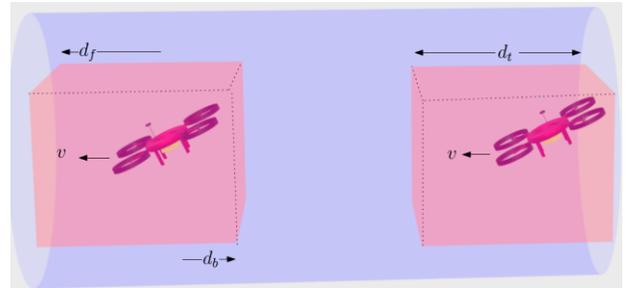

*Fig. 4 UAV placement in core geo-fencing volume to avoid rear-end collisions*

The forward and rear clearances are governed by the speed of the vehicle wherein a higher forward clearance is considered for UAVs moving with faster speeds. The proposed three-layered geo-fencing contains the trajectory deviations within the allocated airspace. In case of violation, the fencing would interpret the UAV motion as a safety breach and issue warning for further actions.

*4. Communication aspects*

The V2V communication is essential for the UAVs while in the corridors. This is primarily to ensure safety of the vehicles in case of imminent collisions and to broadcast relevant local decisions taken by any UAV. V2X communication helps to monitor the vehicles continuously from the base station, so that any off-nominal behavior could be tracked and handled with minimal loss. 5G communication could help in better localization of the drones as relying on GPS alone could lead to erroneous localization.

## 5. Corridor compliance of drones

The vehicles using the corridors require certain basic functionalities. These include both software and hardware aspects. Various levels of compliance could be defined based on the vehicle capabilities, as given in Table 1. The compliance level decides the level of geo-fencing required by a drone and the kind of mission it could be assigned. Having well equipped sensor suite and communication systems aids autonomy. This also implies that the core fencing parameter $d_t$ for CL3 UAVs would be minimal while that for CL1 UAVs would be the maximum. This is explained by the latter's deficiency in hardware and software components.

*Table 1 Compliance levels (CL) for UAVs in CORRIDRONE*

| Level | Capability | Remarks |
|---|---|---|
| CL1 | V2X communication, average localization, guidance and control, short range conflict detection, endurance may be short, no warning module, no fault tolerance. | Low compliance |
| CL2 | Good guidance/control system, mid-range conflict detection module, GCS based health monitoring, V2X communication, GPS based localization, endurance may be short, fencing violation detection and warning, immediate landing on fault. | Medium compliance |
| CL3 | Robust control/guidance system, high range conflict detection module, on-board health monitoring, V2V/V2X communication, accurate localization, endurance for round trip, warning module (fencing violation, hacking attempt, rogue node detection), fault tolerance | High compliance |

Vehicles which are CL3 compliant improve the efficiency and throughput of the architecture. CL1 UAVs could be operated for missions of short duration and distance, in near ideal conditions. In general, those vehicles which support vertical takeoff and landing and have adequate sensing, range, localization, endurance, and telecommunication capabilities are candidate UAVs for corridors. Other specifications of the drones will depend on the utility for which they are employed.

## 6. Miscellaneous aspects

A few other aspects are desirable for the on-demand UAV corridors which are not described in this article. The topographical information is important for design of CORRIDRONES. The temporal aspects related to deployment as well as the safe transit of drones require weather information of that area. Poor localization or GPS denied regions would require alternate methods like vision to execute the mission. Proper propeller designs are required for quieter drone, especially for operation in an urban scenario.

CORRIDRONE would require inputs like starting location and destination, altitude of operation, expected throughput, mission utility, desired flight duration, time of day for mission, etc. The location and altitude details decide the geo-spatial location of the generated corridors. The throughput and utility decides the size of the corridor, lane geometry and drone capability requirements. Given the inputs, the software suit in the GCS would compute the outer fencing parameters, and plan the corridors and lanes. The desired mission time and utility, along with the environmental data would be utilized to compute the velocity bounds. The throughput, compliance level of drone available for mission and lane planning determines the core fencing. The final output from the CORRIDRONE GCS application would provide users with best corridor options. This is forwarded to a higher authority for permission to operate. Based on the allocated volume, a quick generation of drone corridor is carried out to avoid any errors/violations. The duration for which the corridor remains active depends on the mission duration and a calculated buffer time, which is expected to account for any unexpected delays.

## CORRIDRONE as a part of UTM

The details above give the basic idea of what a CORRIDRONE is and its essential features and testbed description. Integration of the proposed drone corridors to the existing traffic architecture is the next important aspect. As mentioned at the beginning, many traffic management architectures are being established for UAVs. They address different aspects of unmanned traffic like the regulatory aspects, air volume management and allocation, routing, traffic monitoring, etc., to name a few. Being a complete module for multi-UAV on-demand corridors, the permissions and airspace allocation are the basic functionalities that connect a UTM and CORRIDRONE (Fig. 5). As discussed earlier, the details of the corridor required for the mission are computed and sent to UTM for approval. The UTM retrieves information of already allocated volumes and fencing to approve the request. Once flight volume gets allocated, the rest of the operations could be executed by the proposed system with minimal dependencies on the traffic management system. This is under the assumption that a central server provides the no-fly zone details of

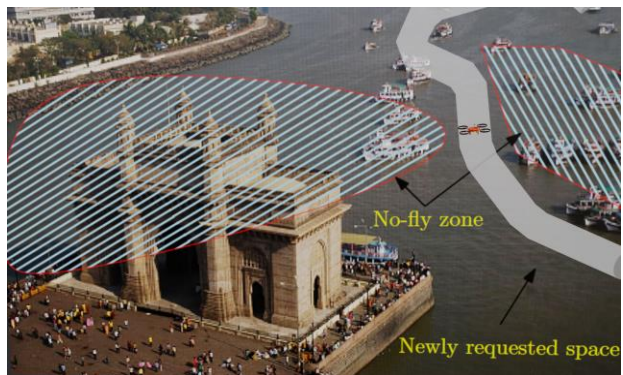

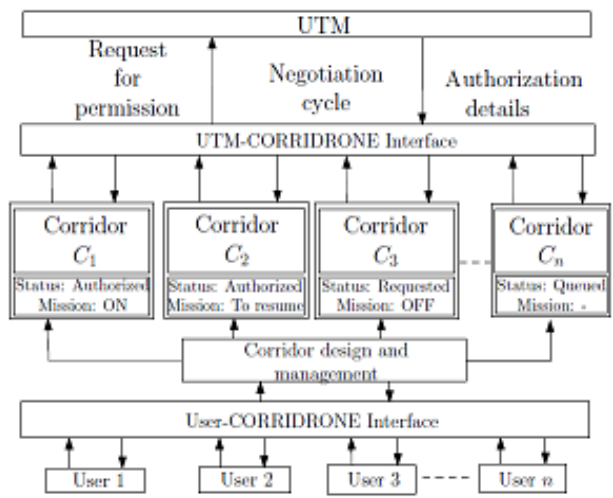

*Fig. 5* (a) Airspace allocation by UTM (b) Integration of UTM and CORRIDRONE

the concerned region and is available for use. Fig. 5(a) shows a possible scenario where the UTM gives an indication of the no-fly zones and already allocated zones during a specific time which the CORRIDRONE operator has requested. The CORRIDRONE GCS now proposes a corridor which is now costed by the UTM and approved. The process can enter further negotiation cycles as indicated in the block schematic (Fig. 5(b)). Communication on the mission completion and release of allocated volume would mark the end of information interchange with UTM.

## Applications of CORRIDRONE

CORRIDRONE finds application in a variety of utilities, a few of which are given below. These are applications that require on-demand setting up of drone corridors rather than permanent ones.

a. *Factory transport utility*
   Main corridor (permanent) and off-corridors (transient, based on demand) designed for multiple deliveries. Avoidance and formation flying techniques could be employed in the main corridor while off corridors branch out/converge to main corridor.

b. *Shore-to-ship delivery utility*
   Conservative corridors for windy environment. Design and corridor life time depending on local climate and duration of mission.

c. *Border patrol utility*
   Flexible corridors in undulating terrain for surveillance; Stealth mode corridors for vigilant operations. See Fig. 6, which shows the CORRIDRONE being set-up on a varying altitudes and flight path scenario.

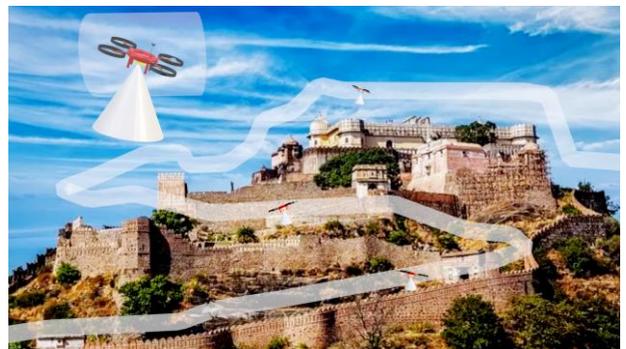

*Fig. 6* Example CORRIDRONE for border patrolling at Kumbhalgarh Fort, Rajasthan

d. *Delivery of essential commodities* (last mile problem)
   Safe corridors for cluttered urban environment

e. *Delivery of urgent commodities* (disaster management/medical emergencies)
   High priority corridors

f. *Precision agriculture*
   Dedicated corridors for monitoring, pesticide spraying (less safety requirements)

## Test-bed and implementation of CORRIDRONE

A test bed for demonstrating the proposed drone corridor is being set up at the Challekere campus of IISc. At the initial stages, the modules of the corridors like adaptive geo-fencing, lane-planning, etc., would be validated. In subsequent phases, corridors would be designed to connect different parts of the campus and the integrated system would be tested. Fig. 7 shows the IISc campus area where the drone corridors set-up is being planned. The first prototype will be set-up in the 2-acre airfield, followed by a 100-acre test bed with a total length of about 3 - 4 km of the corridor arranged as shown in Fig. 8(a). The drone corridor will be initially set up as a set of drone lanes (inward flow, outward flow, service, and emergency) stacked in a 3-D formation that will enable access of the lanes from each of the other lanes.

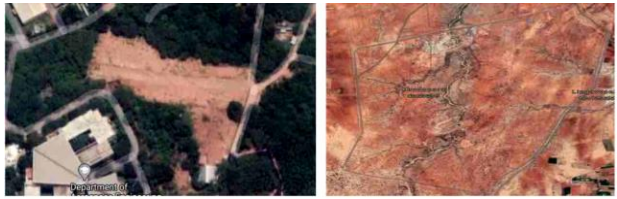

**Fig. 7** *IISc Challekere campus and the campus airfield where test bed for proposed drone corridors are being set up.*

Formation strategy in 3-dimensional space designed to consider the spatial formation of multiple drones, wake and downwash effects, and constraints on the delivery and take-off platforms. The distribution shown in Fig. 8(a) is another possibility of designing lane geometry, besides the ones introduced previously. Each of these have their own advantages which could be chosen based on the requirements of the application. Fig. 8(b) shows the test bed infrastructure schematic, focusing on the communication and network aspects. The cyber-physical system installed at the test rig would enable smooth implementation, and testing of CORRIDRONE and data collection. The flight data opens up possibilities for further research and development of the overall system.

Access to emergency lanes through transfer chutes, and the points where the direction of flight changes, will be designed separately as the typical dynamics of the drones require additional constraints on the lane change maneuver. Monitoring stations at periodic intervals will be used to ensure risk free drone flight through the corridors. Although the flight of the drones is autonomous in certain portions of the flight, it is still restricted by the constraints of corridors space allocated to it, and hence may have limited capability in terms of flight path change. This will constrain its collision avoidance maneuvers and will impact the drone specifications and capabilities that makes the drone "corridor compliant".

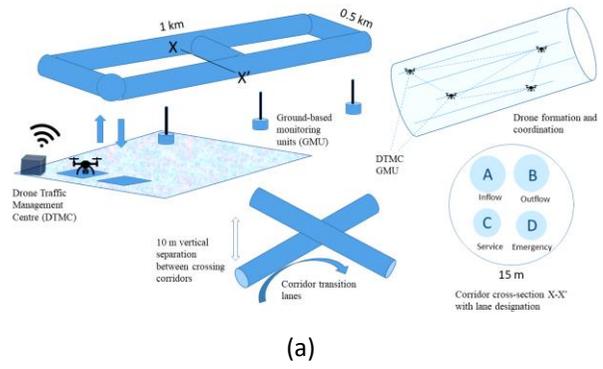

(a)

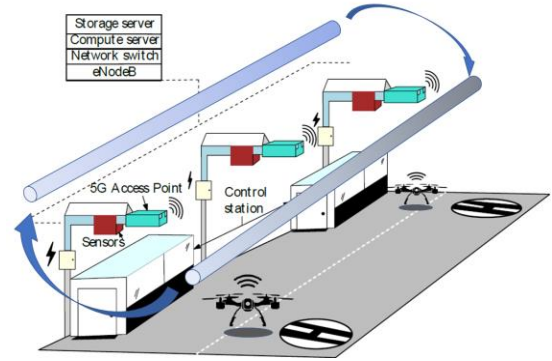

(b)

**Fig. 8** *The CORRIDRONE test-bed and communication infrastructure*

## Conclusions

In this article, a portable drone corridor called CORRIDRONE is proposed. This on-demand air corridor design has many novel features like multi-lane geometry and adaptive dynamic fencing which can be adopted for quick missions with ease. CORRIDRONE also introduces and integrates compliance level of UAVs into the flight plan. The presented module could be integrated with any traffic management system as the dependencies are straightforward. Moreover, a test bed is being set up in the Indian Institute of Science campus for implementing and developing the proposed framework.

## Acknowledgements

Authors wish to acknowledge RBCCPS (IISc) for their funding support. We also acknowledge Prof. B. Amrutur's assistance for the testbed infrastructure sketch, given in Fig. 11, and Stuti Garg, Chaitali Pujar, and Nandni Sharma, undergraduate interns at the Guidance, Control and Decision Systems Laboratory at the Department of Aerospace Engineering, IISc, for their help with the initial stages of literature survey.

## Author Information

**Lima Agnel Tony** is a PhD research scholar in the Department of Aerospace Engineering, Indian Institute of Science, Bangalore, India. Her current research interests are in the development and implementation of algorithms for motion planning and control of autonomous systems, with an emphasis on UAV traffic management. Apart from her PhD research, she is also interested in robotics and had a leading role in the IISc-TCS team which participated in the MBZIRC 2020, involving various challenging applications using drones.

**Ashwini Ratnoo** received his bachelor's degree in Electrical Engineering from the MBM Engineering College, Jodhpur, India, in 2003, and master's and Ph.D. degrees, in Aerospace Engineering, from the Indian Institute of Science, Bangalore, India, in 2005 and 2009, respectively. During 2009-2012, he was a postdoctoral researcher at the Aerospace Engineering Department, Technion- Israel Institute of Technology, Haifa, Israel. Currently, he is an Associate Professor at the Aerospace Engineering Department, Indian Institute of Science, Bangalore, India. His research interests include guidance, path planning, collision/obstacle avoidance, and coordination of unmanned aerial systems. He is an Associate Fellow of AIAA, and a member of AIAA Guidance, Navigation, and Control Technical Committee.

**Debasish Ghose** is a professor in the Department of Aerospace Engineering, at the Indian Institute of Science, Bangalore, India. He is also an adjunct professor with the Robert Bosch Centre for Cyber-Physical Systems (RBCCPS) at IISc, and the PI of the IISc-RBCCPS Drone Skyways project. He has been a visiting faculty at the University of California at Los Angeles in the past. He is in the editorial board of several reputed journals. His areas of interest are guidance and control of autonomous vehicles, multi-agent systems, and robotics. He is an associate fellow of AIAA and a fellow of Indian National Academy of Engineering.